\documentclass{article}

\usepackage[preprint]{neurips_2024}

\usepackage[utf8]{inputenc} % allow utf-8 input
\usepackage[T1]{fontenc}    % use 8-bit T1 fonts
\usepackage{hyperref}       % hyperlinks
\usepackage{url}            % simple URL typesetting
\usepackage{booktabs}       % professional-quality tables
\usepackage{amsfonts}       % blackboard math symbols
\usepackage{nicefrac}       % compact symbols for 1/2, etc.
\usepackage{microtype}      % microtypography
\usepackage{xcolor}         % colors
\usepackage{linguex}

\usepackage{amssymb}  
\usepackage{amsmath}

\usepackage{algorithm}

\usepackage{algpseudocode}

\usepackage{graphicx}

\usepackage{float}

\usepackage{stmaryrd}

\usepackage{color-edits}
\addauthor[SS-T]{shane}{orange}
\addauthor[TM]{tom}{blue}
\addauthor[LY]{leroy}{green}

\title{Minimization of Boolean Complexity in In-Context Concept Learning}

\author{%
  Leroy Z. Wang \\
  Department of Linguistics\\
  University of Washington\\
  \texttt{lryw@uw.edu} \\
  \And
   R. Thomas McCoy \\
     Department of Linguistics\\
  Yale University \\
  \texttt{tom.mccoy@yale.edu} \\
  \AND
  Shane Steinert-Threlkeld \\
  Department of Linguistics\\
    University of Washington\\
  \texttt{shanest@uw.edu} \\
}

\begin{document}

\maketitle

\begin{abstract}

What factors contribute to the relative success and corresponding difficulties of in-context learning for Large Language Models (LLMs)? Drawing on insights from the literature on human concept learning, we test LLMs on carefully designed concept learning tasks, and show that task performance highly correlates with the Boolean complexity of the concept.  This suggests that in-context learning exhibits a learning bias for simplicity in a way similar to humans.
% \tomcomment{I like the abstract! Maybe add one sentence at the end as a higher-level conclusion or takeaway?} 
\end{abstract}

\section{Introduction}\label{intro}

The human conceptual apparatus represents one of the most remarkable aspects of our species' intelligence \citep{big-book-concept}.  In order to understand the ways in which artificial intelligences do and do not resemble our own, understanding their conceptual structure is an important first step.

One prominent tradition argues that concepts are representations in a \emph{language of thought (LoT)} \citep{fodorLanguageThought1975, goodmanConceptsProbabilisticLanguage2015, quilty-dunnBestGameTown2022}.  The recent Bayesian revolution in cognitive science has argued that concept learning exhibits a very strong bias for simplicity: human learners infer the simplest expression in an LoT that is consistent with the data that they have seen \citep{feldmanMinimizationBooleanComplexity2000, chaterSimplicityUnifyingPrinciple2003, goodmanRationalAnalysisRulebased2008, piantadosiLogicalPrimitivesThought2016}.

In this paper, we study \emph{in-context concept learning} with large language models (LLMs), allowing us to address the following questions: when presented with labeled examples of an unknown concept, can an LLM infer the underlying concept?  If so, what inductive biases does this in-context concept learning exhibit; in particular, does it exhibit a simplicity bias akin to the simplicity bias displayed by humans?

Consider the prompt in \ref{ex:alice_3_10}.  In the first two lines, we see labeled examples of a new numerical concept, \emph{bnik}.  The final line asks a model to label a new example.  Repeating this for a range of example sets and concepts, we can measure whether models have greater success with simpler concepts.

\ex. There are 10 apples. Alice has 3 of the apples. Does Alice have bnik of the apples? No.
\\
There are 15 apples. Alice has 10 of the apples. Does Alice have bnik of the apples? Yes.
\\
There are 20 apples.  Alice has 10 of the apples.  Does Alice have bnik of the apples? \_\_\_ \label{ex:alice_3_10}

We study a range of numerical concepts, expressed in a simple language of thought with basic logical and arithmetical operators and find (for several different models) that concepts with shorter representations in a hypothesized LoT are easier to learn in-context.  This shows that LLMs are capable of learning non-trivial mathematical concepts and exhibit learning biases that are similar to those used in human concept learning.

% \shanecomment{I streamlined a very short-and-sweet intro; let me know what you think, and feel free to edit / comment as you see fit!  I tried to add some emphasis of numerical/methematical concepts/reasoning here at the end for the venue} \tomcomment{Looks great to me! I like adding that math-y spin.}

% Consider the labeled examples above (first 4 lines) \tomcomment{I like using the package linguex for examples because then you can refer to them - I added a linguex example for the first one such that it can be referred to as example \ref{ex:alice_3_10}}, where the nonce word \textit{bnik} represents the underlying concept of \textit{between 5 and 10}. In other words,  
% 
% \vspace{-0.3cm}
% 
% \begin{align*} 
% & \textnormal{``the semantics of \textit{bnik}''}   \\
% &= \llbracket \textnormal{bnik} \rrbracket \\
% &= \llbracket \textnormal{between 5 and 10} \rrbracket \\
% &=   \{ \langle n, x \rangle : (x > 5) \wedge (10 > x) \},
% \end{align*}
% 
% where $n$ represents the total number of items, and $x$ represents the number of items of interest.
% \shanecomment{We don't need to use this many lines to state this, and also $n$ and $x$ don't correspond to the example above.}
% 
% 
% The LM would be provided with a set amount of labeled examples, and an unlabeled one at the end, where the LM will be prompted to label it.
% \shanecomment{Summary paragraph here}

\section{Related Work}

\citet{feldmanMinimizationBooleanComplexity2000} showed that ease of human concept learning is highly negatively correlated with Boolean logical complexity: concepts with longer minimal logical formulas are harder for people to learn.  A large body of subsequent work has extended the range and scope of this view using Bayesian inference in various LoTs \citep{goodmanRationalAnalysisRulebased2008, piantadosiLogicalPrimitivesThought2016}. \citet{neural-net-track-complexity} show that neural networks trained from scratch to learn Boolean concepts exhibit a similar bias for simplicity.  A wide range of work has recently analyzed when, how, and why in-context learning (ICL) in LLMs works \citep[i.a.]{min-etal-2022-rethinking, akyurekWhatLearningAlgorithm2022, pmlr-v235-akyurek24a}. To the best of our knowledge, ours is the first to explicitly study concept learning and measure a learning bias for logical simplicity in ICL.
% \shanecomment{LMK your thoughts on this short-and-sweet section too.  It was especially hard to know what to cite for `analyzing ICL', since that's such a huge field; I don't know of a survey paper, so I went for a few highlights, but if you do, that would be great.} \tomcomment{This looks great too! I'm not coming up with a survey paper either.}

% In this work, we focus on numerical (Boolean) concepts, since their complexity can be precisely defined and measured. Human Boolean concept acquisition has been extensively studied in cognitive science. One of the most influential theories is the language of thought hypothesis (LOTH) \citep{lang-of-thought}, which suggests that human thinking process can be explained by symbol manipulations of a mental language. Some of the more recent works have used language of thought to study the difficulty / complexity of concept acquisition. The seminal work by \cite{human-minimization-bool} showed that human concept learning performance can be predicted by the Boolean complexity of the concept. \cite{rational-analysis-rule} demonstrated that similar phenomena exist in Bayesian learning models. For neural networks, \cite{neural-net-track-complexity} found that neural networks take a longer time to learn categorical concepts with higher Boolean complexities.  

\section{Methodology}

\subsection{Concept generation}
Our data generation methodology is inspired by \citet{quant-min-descr-length} and \citet{gqg}, where we define the complexity of a concept using its minimal description length---the length of the shortest expression that can capture the concept (defined more precisely below). We define a concept as a semantic object generated by a logical grammar, whose basic structure is shown in Table~\ref{operators}. The full grammar used during generation is given in Appendix~\ref{full-grammar}, which imposes some additional constraints to prevent certain types of unwanted recursive generation.\footnote{See Appendix~\ref{full-grammar} for an example.} Code for generating the concepts, as well as prompting models and analyzing data, may be found at \url{https://github.com/lerow/llm-concept-learning-complexity}.

%\tomcomment{Seeing the grammar laid out like this is very helpful! Maybe this is my linguist upbringing talking, but I think the grammar might be easier to understand if we write it as a CFG? With rules like "C $\rightarrow$ Num = Num" and "C $\rightarrow$ C $\wedge$ C" and "Num $\rightarrow$ 7" and "Num $\rightarrow$ x". And then might also be helpful to provide one brief derivation - "Start with start symbol $C$, then use rule 2 to expand it to $C \wedge C$, then use rules 7 and 11 to expand that to $Num > Num \vee Num = Num$, then use terminal rules to expand that to $n > 7 \vee n = 4$. We might be able to have the grammar and an example derivation side by side, to save space?}
% \leroycomment{I'm leaning toward not modifying the grammar, since - there are other pieces of writing that are based on operator count, so having a table of operators seems intuitive - the full grammar for generation is given in appendix, we can also point readers to that} \tomcomment{I'm good with this!} \shanecomment{Same!  We can also revisit this for the full paper in the future; I have lots to say about the operator vs CFG view, which are very intimately related}

\begin{table}[ht]
\centering

\begin{tabular}{ lll } 
\toprule
\textbf{Operator} & \textbf{Type} & \textbf{Gloss}  \\
\midrule

$=$ & {\small NUM $\times$ NUM $\rightarrow$ BOOL} & Numerical equality\\ 

$\neq$ & {\small NUM $\times$ NUM $\rightarrow$ BOOL} & Numerical inequality\\

$>$ & {\small NUM $\times$ NUM $\rightarrow$ BOOL} & Numerical more than\\ 

$<$ & {\small NUM $\times$ NUM $\rightarrow$ BOOL} & Numerical less than\\ 

$\times$ & {\small NUM $\times$ NUM $\rightarrow$ NUM} & Numerical multiplication\\ 

$\wedge$ & {\small BOOL $\times$ BOOL $\rightarrow$ BOOL} & And \\ 

$\vee$ &  {\small BOOL $\times$ BOOL $\rightarrow$ BOOL} & Or\\ 
\bottomrule
\end{tabular}

\caption{Operators in the logical grammar.}
\label{operators}
\end{table}

The complexity of a concept is determined by the number of operators in its minimal description. For example, the concept $\llbracket \textnormal{between 5 and 10} \rrbracket$ has a complexity of 3 (and therefore will be in \textit{class 3}) because there are three operators ($>$, $\land$, and $>$) in its minimal description: $(x > 5) \land (10 > x)$. One example of a concept with a complexity of 1 is $\llbracket \textnormal{less than 5} \rrbracket$, with minimal description: $(x < 5)$.
Concept complexity class $n$ will thus contain all unique concepts with minimal description lengths of $n$ that can be generated by the logical grammar. The complexity classes, along with some representative concepts for each class, are included in Appendix ~\ref{complexity-classes}. 
% We use the ULTK library\footnote{\url{https://github.com/CLMBRs/ultk}} to find all unique concept expressions in each class, and generate all possible expressions where the total number of objects is between $[5, 100]$. 

Under the procedure described so far, it is possible to generate two concepts that have similar or identical meanings.  Here, concepts classify pairs of numbers (in \ref{ex:alice_3_10}, the number of apples and the number that Alice has) into true and false.  We identify a concept's meaning as its extension, i.e. the set of the set of such pairs that it maps to true. 

% \tomcomment{Is there a better definition of the extension than "set of situations"?}

This fact creates several issues for interpreting the results:
\begin{enumerate}
    \item If two concept descriptions in different complexity classes yield the same meaning (e.g., $x < 5 \wedge x < 6$ and $x < 5$), then the more complex one should not be considered because our hypotheses are about a concept's \textit{minimal} description length; thus, only the simplest description for a concept should be considered.
    \item If two concept descriptions in the same complexity class yield the same meaning (e.g., $x < 5 \vee x > 17$ and $x > 17 \vee x < 5$), then they are effectively the same concept. Thus, generating both concepts is effectively the same as sampling one concept twice, which is undesirable because it might make that concept overrepresented, biasing the results.
    \item If two concept descriptions yield similar but non-identical meanings, there are potential challenges in interpreting a model's performance. Consider the concepts with meanings $ (x > 5)$ and  $(x > 5) \wedge (x \neq 7) $. When $n=100$, the second meaning only differs from the first at exactly one place (when $x = 7$). Thus, if we intend to test learning of the second, more complex concept, a model would get almost the same accuracy if it had learned the incorrect simpler concept as it would get if it had learned the correct concept, since both concepts almost always yield identical predictions. This is a problem since it means we cannot tell if the model has learned the intended concept (as opposed to a similar but unintended concept).
\end{enumerate}
To address these issues, we perform deduplication in which a generated concept is discarded if its meaning is the same as, or similar to, a previously-generated concept. Since concepts are generated in order of increasing complexity, this procedure ensures that we keep only the minimal description for a given meaning. For deduplication across complexity classes, we consider two concepts similar when their Levenshtein distance\footnote{See Appendix~\ref{dedup-details} on how the distance is calculated.} is less than $3$, i.e.\ they differ in truth value on at most 3 inputs, in which case we discard the one with the longer description. Deduplication is also performed within complexity classes: when multiple concepts in the same class have a Levenshtein distance of $0$ with each other, we discard all but one of them.

\subsection{Data generation} 
 % \tomcomment{I think the info in this mini paragraph would make more sense in the next subsection? Talking about MDL and complexity might not make much sense until we get to that part.} 
 
 % \tomcomment{I would lean toward swapping this subsection (data generation) with the next (concept complexity classes, except maybe renamed "concept generation"? But I am good with the current order if you prefer}

Each prompt that we give to LLMs, such as the prompt shown in  \ref{ex:alice_3_10} above, is made of several examples. Each example is generated from the following template, where the slots that vary between examples are underlined:

\ex. There are  $\underset{\text{TOTAL}}{\underline{\hspace{1cm}}}$ \, $\underset{\text{OBJ}}{\underline{\hspace{1cm}}}$. \;  $\underset{\text{SUBJ}}{\underline{\hspace{1cm}}}$ \, $\underset{\text{PRED}}{\underline{\hspace{1cm}}}$ \, $\underset{\text{NUM}}{\underline{\hspace{1cm}}}$ of the  $\underset{\text{OBJ}}{\underline{\hspace{1cm}}}$.\\ 
\\
Does $\underset{\text{SUBJ}}{\underline{\hspace{1cm}}}$ \, $\underset{\text{PRED}}{\underline{\hspace{1cm}}}$ bnik of the $\underset{\text{OBJ}}{\underline{\hspace{1cm}}}$? \, $\underset{\text{YES/NO}}{\underline{\hspace{1cm}}}$.   \label{ex:template}

% \leroycomment{yay LaTex magic!} \tomcomment{Beautiful!} \shanecomment{chef's kiss!}

%\tomcomment{Below this, insert a depiction of the template (I think it will be more understandable if there is something concrete that readers can see). The template could be "There are \underline{15} \underline{apples}. \underline{Alice} has \underline{10} of the \underline{apples}. Does \underline{Alice} have \underline{bnik} of the \underline{apples}? \underline{Yes}". Or, instead of filling the slots with actual words (like "Alice" etc.), could give them names like "NAME", "TOTAL", etc. Or with some fancy LaTex-ing, might be able to write "Alice" etc. with an underline, then a generic term like NAME under the underline, Mad-Libs-style?}

%A template is used to generate the prompts in the dataset. 
As shown in Algorithm~\ref{algo-datagen} in Appendix~\ref{app:datagen}, we iterate through all meaningful numerical ranges for both the number of total objects (which we restrict to be between 5 and 100 inclusive) and the number of objects the person has, and generate an example for each combination. If we want to generate a prompt with $m$ positive examples and $n$ negative examples, we would sample without replacement $m$ and $n$ examples from the sets of positive examples and negative examples respectively, and use an unseen example as a question at the end. The sets of examples used in this paper are balanced -- in every prompt, the model sees the same number ($10$) of positive and negative examples\footnote{Because of this, we exclude concepts that would have fewer than 10 positive or negative examples.}; and for each accuracy data point, the model is tested on the same number ($500$) of prompts with true labels and false labels.  In this work, we use the template ``\textit{Let us define a new word, bnik}.'', followed by labeled examples and a question at the end, for all prompts in the dataset.

% \tomcomment{I propose adding this parenthetical here and cutting "where the total number of objects is between [5, 100]" from Section 3.1, since I think it feels a little out of place in 3.1. I don't feel strongly though. We should only keep it in one place, so either delete this new one or delete the one in Section 3.1}

\subsection{Models}

We ran experiments on two LLM families: Qwen2 \citep{qwen2-report} from Alibaba research, and Gemma 2 \citep{gemma2-report} from Google DeepMind. During testing, the instruction-tuned versions of the models and default Huggingface chat templates were used. Qwen2-72b was the best-performing open model on Hugging Face Open LLM Leaderboard\footnote{\url{https://huggingface.co/spaces/open-llm-leaderboard/open_llm_leaderboard}} as of June 2024. The Gemma 2 models achieved state-of-the-art results for their size, while reaching competitive performance on many benchmarks when compared to models with $2 \times$ more parameters. 

% \tomcomment{Can you give a link or citation to the recommended Huggingface chat templates? Or are they specified in the papers that are already cited?}

\section{Results}

\begin{figure}[H]
    \centering
    
\includegraphics[scale=0.22]{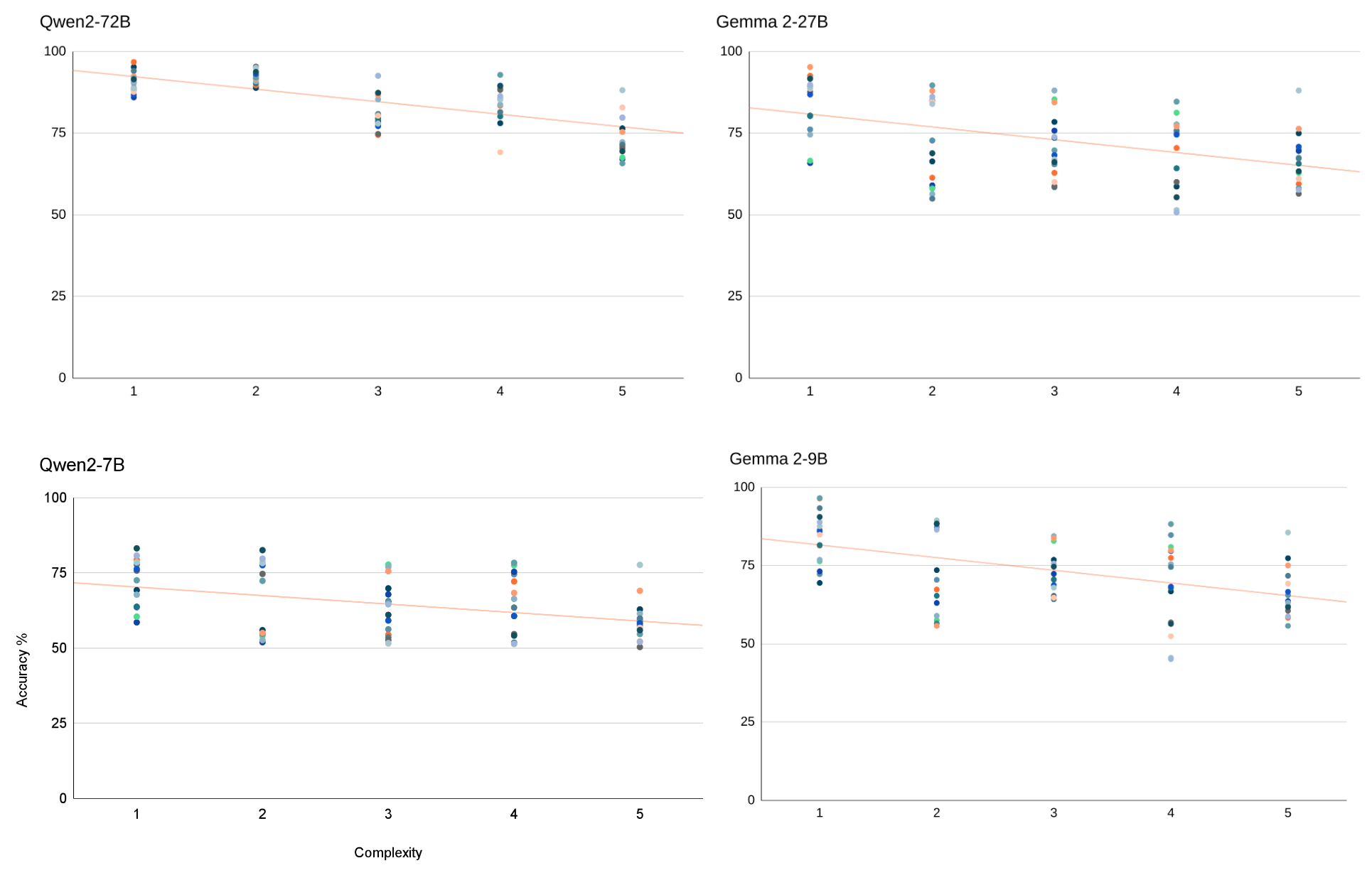}

    \caption{The influence of concept complexity on LLM accuracy at concept learning (see text for how complexity is operationalized). On average, LLM accuracy drops as complexity increases.}
 \label{fig:main-result}
    
\end{figure}

For each complexity class, we randomly sample $18$ concepts from all possible concepts in that class. Each data point on the plot in Figure \ref{fig:main-result} represents the model's accuracy on a specific concept. The trend line shows the line of best fit to the average accuracy of each complexity class. 

% \tomcomment{I think this last sentence isn't quite accurate (unless you're planning to modify the plot?) E.g., in complexity class 2 for Qwen2-72B, it's clear that the trend line is not going through the average value. So I think it's plotting something other than average accuracy for each class, such as a linear regression fitted to the entire dataset?}

% \leroycomment{Hmmm, on Google sheets it says it's plotting just the average values. I think it's just a line of best fit?} \tomcomment{That would make sense! In that case I think we should say "the trend line shows a line of best fit to the average accuracy for each complexity class."}

For each model family, we run experiments for models in two sizes -- the largest model in that family, and a model that has approximately 10 billion parameters. 
As shown in the figure, the average accuracy for all LLMs decreases as concept complexity increases. The drop in average accuracy is most evident in Gemma 2-9B: from $83\%$ in class 1 to $66\%$ in class 5. For the largest model we tested, Qwen2-72B, there is a $16\%$ decrease ($90\% \rightarrow 74\%$) in average accuracy as complexity increases from 1 to 5. See Appendix ~\ref{avg-plot} for a plot of only average accuracy values for each model.

In Table~\ref{table:pcc}, we see there is a strong negative correlation between concept complexity and average model accuracy, indicated by the Pearson correlation coefficients (PCC). All results are statistically significant except for Qwen2-72B, which is nearly significant at a $p = 0.05$ threshold.

% [** This plot above will be improved before submission **]

% \leroycomment{I think I'll put an additional plot of just average values, either here or in appendix depending on space}

% \shanecomment{Can you write some prose walking the reader through the Figure and the Table, so that this reads like a complete section?  You can that they are all statistically significant (which test is used? read the docs for whatever software you used for that) except for Qwen 72B, which is ``nearly significant at a $p=0.5$ threshold''}

% \shanecomment{The results section (either a sentence in the last paragraph or a new one) should make explicit reference to the table of correlations and tell the reader what they mean.}

\begin{table}[H]
    \centering
    \begin{tabular}{lcc}
        \toprule
        \textbf{Model Name} & \textbf{PCC} & \textbf{$p$-value} \\
        \midrule
        Gemma-2-\textbf{9B}-it &  -0.961 & 0.009 \\
        Gemma-2-\textbf{27B}-it & -0.898 & 0.038 \\
        Qwen2-\textbf{7B}-Instruct & -0.884 & 0.046 \\
        Qwen2-\textbf{72B}-Instruct & -0.854 & 0.065 \\
        \bottomrule
    \end{tabular}
    \caption{Pearson correlation between complexity and average accuracy}
    \label{table:pcc}
\end{table}

% \shanecomment{Are you adding Aya results and a figure?  I thought at our last meeting we established that all the results were in, i.e.\ just the four models.  Let us know if/when we can expect more (though I am very OK with not including aya at this stage)!}

\section{Conclusion}

This work has shown that LLM in-context concept learning exhibits a simplicity bias of a similar sort as the simplicity bias that has been observed in human concept learning.  Much work remains for the future: (i) more detailed comparisons with human concept learning data (e.g., of exact learning curves), (ii) extending this research to conceptual domains beyond the numerical, and %\tomcomment{I like the additional subtle reinforcement of ``Yes, this paper is about numerical reasoning!''}, 
(iii) more detailed analysis of factors explaining the ease of in-context concept learning beyond LoT complexity.  We hope that this work will spur a new line of research exploring how perspectives from the concept learning literature can shed light on the nature of in-context learning in LLMs. % \tomcomment{I thought we could have something stronger for this final line (previously said "We hope that this will spur a new line of research on in-context concept learning".) So I suggested what's here now but am not particularly attached to it.} 

% \shanecomment{Foreshadowing our long paper title :-p}

%This enables more research into LLM's similarity with human behavioral data. **cite more than vs. less than Grodzinsky paper

%One possible direction of future work is to extend the domain of concepts to more diverse categories beyond numerical meanings.

%%%%%%%%%%%%%%%%%%%%%%%%%%%%%%%%%%%%%%%%%%%%%%%%%%%%%%%%%%%%

% \shanecomment{Leroy, a note on the references: I added some of my own and may have introduced some duplicates (e.g. Feldmanand Goodman et al 2008). (1) Can you make sure all references are to the same version (I think they are now)? (2) For remaining citations that you got from Semantic Scholar: can you chase down fully accurate citation information, especially when it comes to URLs?  It's a bit ludicrous that SS adds links to their site in the BibTeX instead of the genuine links, but we need the latter for sound scholarship.}

% \leroycomment{ Fixed all citation issues, I think}

\bibliography{citation}

\appendix

\section{Appendix / supplemental material}

\subsection{Limitations}

Only a finite set of fractions are used in the grammar to improve efficiency. As a consequence, it may be the case that some concepts that require a certain minimal number of operators under our framing could be expressed using fewer operators if more fractions were allowed.. To address this issue, we plan to use a richer set of fractions in future work.

Only a limited set of LLMs are tested. It is possible that newer models / models with different architectures do not exhibit the phenomena discussed in this text.

We use only one prompt template for all experiments, which may introduce implicit biases in the data and affect the experiment results.

\subsection{Complexity vs. average accuracy plot} \label{avg-plot}

\begin{figure}[H]
    \centering
    
\includegraphics[scale=0.22]{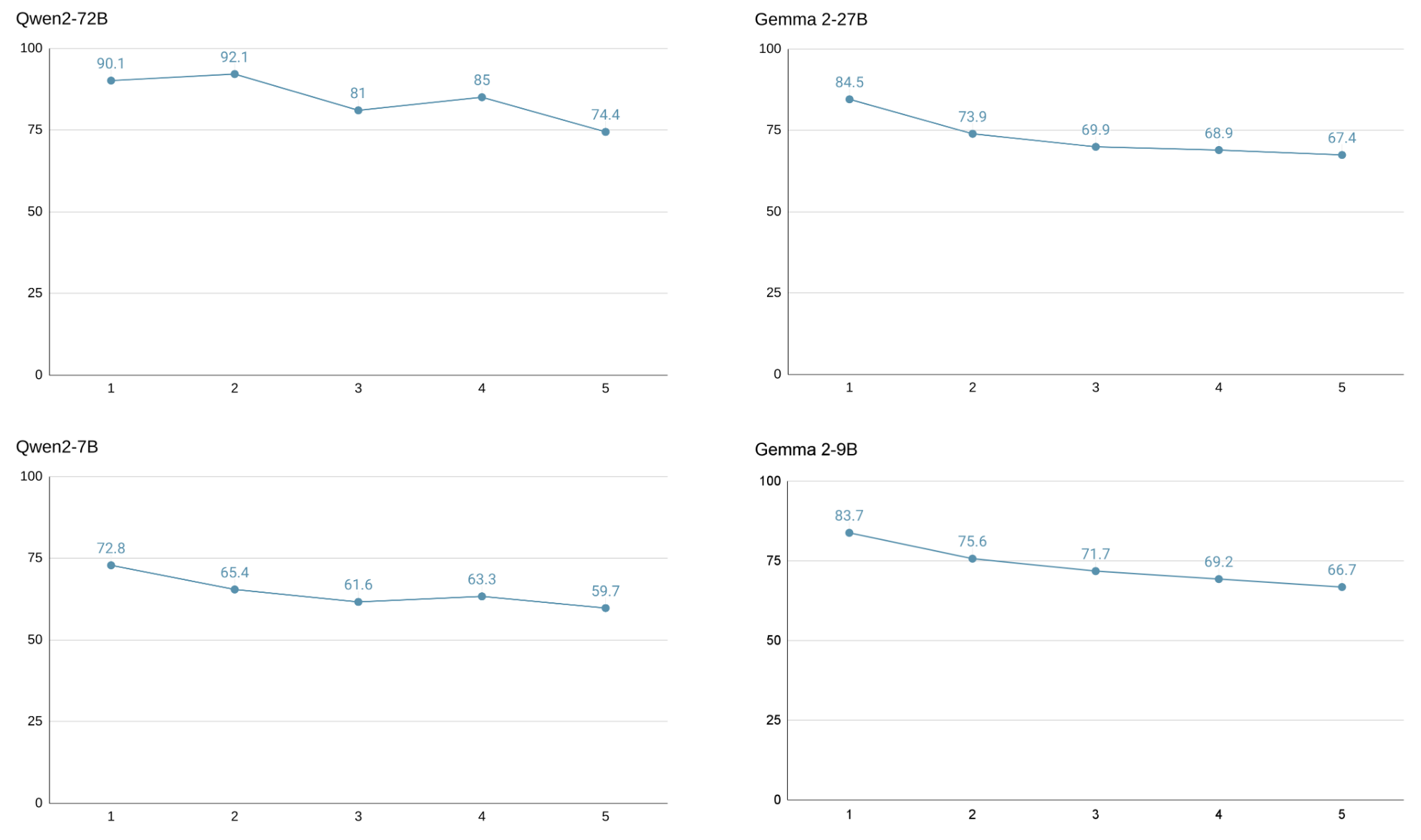}

    \caption{Complexity vs. average accuracy.}
 \label{fig:main-result-avg}
    
\end{figure}

\subsection{Data Generation}\label{app:datagen}

The pseudocode shows how prompt data is generated, from sets of (predefined) subjects, verbs, and objects.

\begin{algorithm}[H]

\caption{Data Generation}
\label{algo-datagen}

\begin{algorithmic}

\State \textbf{Inputs}: set of subject nouns $S$, set of predicate verbs $P$, set of objects $O$,\\ function  {\small GENERATE\_EXAMPLE} \\ \\

initialize $positive\_examples = [ ]$, $negative\_examples = [ ]$ \\

\For{$total$ \, in $[5, 100]$}
\For{$num$ \, in $[0, $ total$]$}

\State uniformly randomly sample $s, p, o$ from $S, P, O$
\State $example$ = {\small GENERATE\_EXAMPLE} $(total, num, s, p ,o)$ \\
\State append example with true labels to $positive\_examples$
\State append example with false labels to $negative\_examples $

\EndFor
\EndFor \\

\State \textbf{return} $positive\_examples, negative\_examples$

\end{algorithmic}

\end{algorithm}

\subsection{Grammar for concept generation}
\label{full-grammar}

The grammar used during concept generation is given below, presented as a context-free grammar.

\begin{verbatim}

Bool -> Bool   and   Bool
Bool -> Bool   or    Bool

Bool -> Var2   ==    SimpleInt
Bool -> Var2   !=    SimpleInt
Bool -> Var2   >     SimpleInt
Bool -> Var2   <     SimpleInt

Bool -> Var2   >     ComplexFloat
Bool -> Var2   <     ComplexFloat


ComplexFloat -> SimpleFloat   *   Var1


# total number of items
Var1 -> [5, 100]

# subject's number of items
Var2 -> [0, 100]

SimpleInt -> [0, 100]   

# fractions
SimpleFloat -> {1/5, 1/4, 1/3, 2/5, 1/2, 3/5, 2/3, 3/4, 4/5}

\end{verbatim}

\paragraph{Constraints on the grammar} For the sake of efficiency, the fine-grained type system in the grammar prevents the generation of fractions that are not in the predefined list of fractions. Multiplication can only take place between "SimpleFloat" and "Var1", so it is not possible to obtain new fractions by multiplying existing ones.

\subsection{Deduplication method} \label{dedup-details}

The deduplication methodology used here is based on vectors representing concept meanings. Suppose the total number of objects is $n=100$. We can then imagine the semantics of a concept being represented by a 101-dimensional vector, where each dimension is the truth value of $f(n=100, x) = \{ \langle n, x \rangle : (x > 3)\}$ for $x = [0, 100]$. 

For instance,

\begin{verbatim}
    x=0    (x > 3) False    0
    x=1    (x > 3) False    0
    x=2    (x > 3) False    0
    x=3    (x > 3) False    0
    x=4    (x > 3) True     1
    x=5    (x > 3) True     1
    ......
    
    -> [0 0 0 0 1 1 ......].
\end{verbatim}

We compute such a semantics vector for $n=\{25, 50, 100\}$ for all concepts in each class, and use the edit distance\footnote{since any two concept vectors always have the same length, the distance is defined as the number of places where two vectors have different values.} to remove similar concepts. Two concepts are considered similar if they belong to different classes and have an edit distance $< 3$ between their vectors. 

% \tomcomment{How is Levenshtein distance defined here? Each insertion/substitution/deletion counts as 1 edit?} 

\begin{algorithm}

\caption{Concept deduplication: when two concepts are similar, only remove the one in the more complex class}
\label{algo-dedup}

\begin{algorithmic}

\State \textbf{Inputs}: current concept class $this\_class$, set of all previous concept classes $prev\_classes$\\ function  {\small EDIT\_DIST} \\

\State $deduped\_concepts = this\_class$ \\

\For{$concept$ \, in $this\_class$}
\For{$prev\_concept$ \, in $prev\_classes$}

    \If{{\small EDIT\_DIST} ($concept$, $prev\_concept$) $< 3$}
        \State remove $concept$ from $deduped\_concepts$
    \EndIf

\EndFor
\EndFor \\

\State \textbf{return} $deduped\_concepts$

\end{algorithmic}

\end{algorithm}

\subsection{Example prompt}

\begin{table}[H]
\centering

\begin{tabular}{ l|l } 
\hline
\textbf{Prompt} & \textbf{Label} \\
\hline

Let us define a new word, bnik. & \\
There are 17 plants. Alice has 13 of the plants. & \\
Does Alice have bnik of the plants? No. & \\
There are 99 trees. Bob has 7 of the trees. & \\
Does Bob have bnik of the trees? Yes. & \\
There are 40 tables. Alice owns 36 of the tables. & \\
Does Alice own bnik of the tables? No. & \\
There are 72 chairs. Bob owns 9 of the chairs. & \\
Does Bob own bnik of the chairs? Yes. & \\
There are 82 chairs. Bob has 47 of the chairs. & \\
Does Bob have bnik of the chairs? No. & \\
There are 100 plants. Alice owns 70 of the plants. & \\
Does Alice own bnik of the plants? No. & \\
There are 56 chairs. Alice owns 3 of the chairs. & \\
Does Alice own bnik of the chairs? Yes. & \\
There are 56 birds. Bob owns 37 of the birds. & \\
Does Bob own bnik of the birds? No. & \\
There are 84 tables. Alice owns 12 of the tables. & \\
Does Alice own bnik of the tables? Yes. & \\
There are 69 bikes. Alice owns 32 of the bikes. & \\
Does Alice own bnik of the bikes? Yes. & \\
\textbf{There are 99 apples. Alice has 3 of the apples.} & \\
\textbf{Does Alice have bnik of the apples?} & Yes \\ 

\hline

\end{tabular}

\caption{Example of prompts in the dataset. The underlying concept in this example is ``less than half''.}
\label{table-prompt}
\end{table}

\paragraph{Lexical items used in prompts}
\begin{itemize}
    \item nonce word: \textit{bnik}
    \item subjects: Alice, Bob
    \item predicates: has, owns
    \item objects: "tables", "chairs", "apples", "bikes",
               "trees", "fish", "birds", "plants"
\end{itemize}

\subsection{Compute resources}
All experiments were run on two Nvidia RTX A6000 GPUs. Greedy decoding was used during inference.

\newpage
\subsection{Concept complexity classes}  \label{complexity-classes}

Below are some representative concepts for each class. It is not a comprehensive list of all the concepts that are generated.

\begin{figure}[H]
    \centering
    
\includegraphics[scale=0.5]{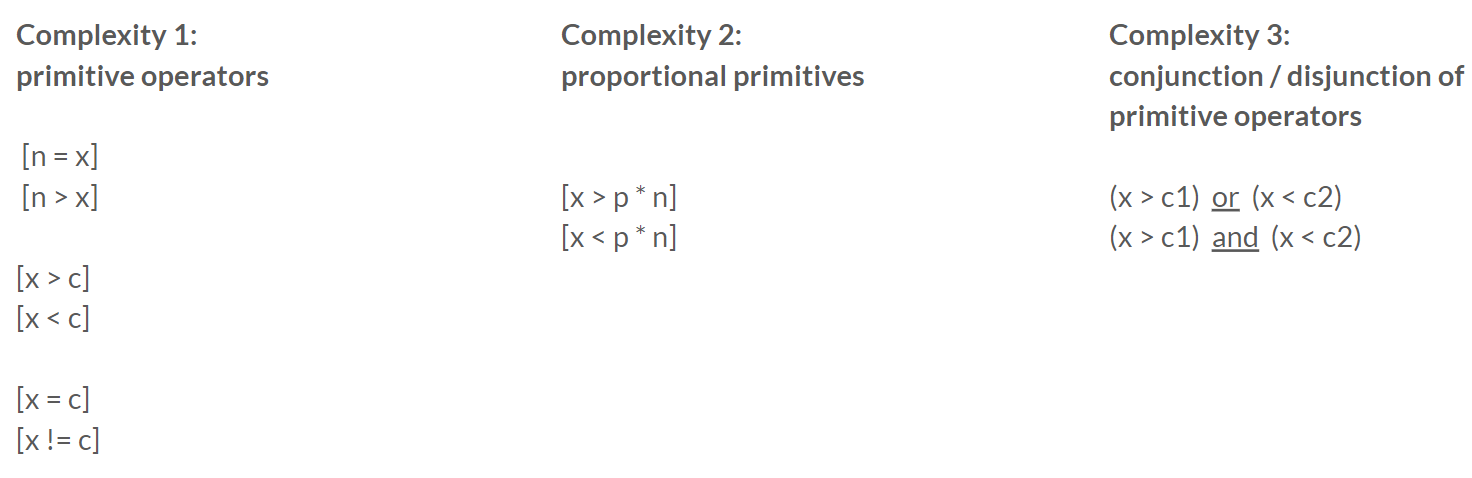}

    \label{fig:c1}
    
\end{figure}

\begin{figure}[H]
    \centering
    
\includegraphics[scale=0.5]{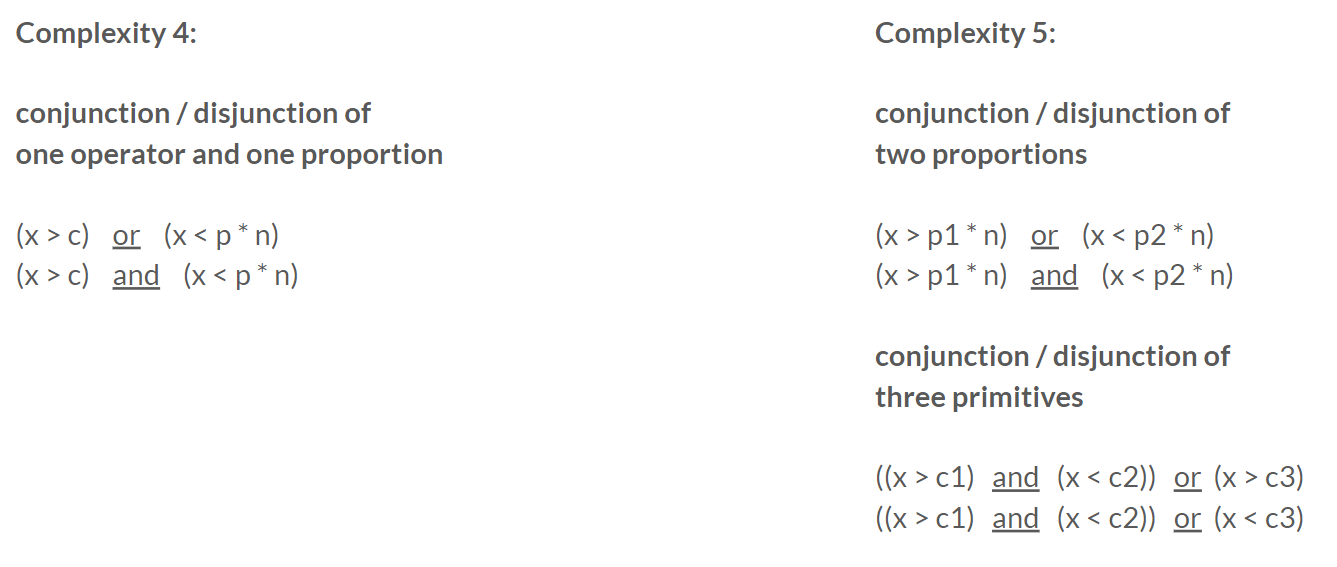}

    \label{fig:c2}
    
\end{figure}

\end{document}